\documentclass[journal]{IEEEtran}
\usepackage{float}
\usepackage{amsmath,amsfonts}
\usepackage{amssymb} 
\usepackage{algorithmic}
\usepackage{array}
\usepackage[caption=false,font=normalsize,labelfont=sf,textfont=sf]{subfig}
\usepackage{textcomp}
\usepackage{stfloats}
\usepackage{url}
\usepackage{verbatim}
\usepackage{graphicx}

\usepackage{tabularx}
\usepackage{longtable}
\usepackage{booktabs}
\usepackage{colortbl}
\usepackage{enumerate}
\usepackage{multirow}
\usepackage{multicol}

\usepackage{algorithm}
\usepackage{xcolor}
\usepackage{amsmath, amsfonts, amsthm, stackrel, amssymb} 
\usepackage{float}
\usepackage{comment}
\usepackage{color}
\usepackage{makecell}
\usepackage{subcaption}
\usepackage{float}
\usepackage{bbding}
 
\usepackage{newfloat}
\usepackage{listings}
\newcommand{\etc}{\textit{etc}.}
\newcommand{\ie}{\textit{i}.\textit{e}.}
\newcommand{\eg}{\textit{e}.\textit{g}.}

\usepackage[colorlinks,linkcolor=black, anchorcolor=black, citecolor=black]{hyperref}

\def\BibTeX{{\rm B\kern-.05em{\sc i\kern-.025em b}\kern-.08em
    T\kern-.1667em\lower.7ex\hbox{E}\kern-.125emX}}
\usepackage{balance}

\usepackage{fancyhdr}

\begin{document}

\title{Finger in Camera Speaks Everything: Unconstrained Air-Writing for Real-World}

\author{Meiqi Wu \IEEEmembership{Member, IEEE}, Kaiqi Huang \IEEEmembership{Senior Member, IEEE}, Yuanqiang Cai, Shiyu Hu, Yuzhong Zhao \IEEEmembership{Member, IEEE}, Weiqiang Wang* \IEEEmembership{Member, IEEE}

\thanks{Meiqi Wu, Yuzhong Zhao, Weiqiang Wang are with the
School of Computer Science and Technology, University of Chinese Academy
of Sciences, Beijing 100049, China, E-mail: {wumeiqi18, zhaoyuzhong20}@mails.ucas.ac.cn, wqwang@ucas.ac.cn.}

\thanks{Yuanqiang Cai is currently a lecturer at the Beijing University of Posts and Telecommunications, Beijing 100876, China, E-mail: caiyuanqiang15@mails.ucas.ac.cn.}

\thanks{Shiyu Hu is with the School of Artificial Intelligence, University of Chinese Academy of Sciences, Beijing 100049, China, and also with the Center
for Research on Intelligent System and Engineering, Institute of Automation, Chinese Academy of Sciences, Beijing 100190, China.
E-mail: hushiyu2019@ia.ac.cn.}

\thanks{Kaiqi Huang is with the Center for Research on Intelligent System and Engineering and National Laboratory of Pattern Recognition, Institute of Automation, Chinese Academy of Sciences, Beijing 100190, China, and with the
University of Chinese Academy of Sciences, Beijing 100049, China, and also
with the CAS Center for Excellence in Brain Science and Intelligence Technology, Shanghai 200031, China. E-mail: kqhuang@nlpr.ia.ac.cn.}


}

\markboth{IEEE TRANSACTIONS ON CIRCUITS AND SYSTEMS FOR VIDEO TECHNOLOGY, VOL. X, NO. X, X X}%
{How to Use the IEEEtran \LaTeX \ Templates}

\maketitle

\IEEEpubid{\begin{minipage}{\textwidth}\ \\[30pt] \centering
		Copyright \copyright 2024 IEEE. Personal use of this material is permitted. 
		However, permission to use this material for any other purposes must \\ be obtained 
		from the IEEE by sending an email to pubs-permissions@ieee.org.
\end{minipage}}


\begin{abstract}
Air-writing is a challenging task that combines the fields of computer vision and natural language processing, offering an intuitive and natural approach for human-computer interaction.
However, current air-writing solutions face two primary challenges: (1) their dependency on complex sensors ($e.g.$, Radar, EEGs and others) for capturing precise handwritten trajectories, and (2) the absence of a video-based air-writing dataset that covers a comprehensive vocabulary range. These limitations impede their practicality in various real-world scenarios, including the use on devices like iPhones and laptops.
To tackle these challenges, we present the groundbreaking air-writing Chinese character video dataset (AWCV-100K-UCAS2024), serving as a pioneering benchmark for video-based air-writing. This dataset captures handwritten trajectories in various real-world scenarios using commonly accessible RGB cameras, eliminating the need for complex sensors. AWCV-100K-UCAS2024 includes 8.8 million video frames, encompassing the complete set of 3,755 characters from the GB2312-80 level-1 set (GB1).
Furthermore, we introduce our baseline approach, the video-based character recognizer (VCRec). VCRec adeptly extracts fingertip features from sparse visual cues and employs a spatio-temporal sequence module for analysis. Experimental results showcase the superior performance of VCRec compared to existing models in recognizing air-written characters, both quantitatively and qualitatively. This breakthrough paves the way for enhanced human-computer interaction in real-world contexts. Moreover, our approach leverages affordable RGB cameras, enabling its applicability in a diverse range of scenarios. The code and data examples will be made public at https://github.com/wmeiqi/AWCV.
\end{abstract}

\begin{IEEEkeywords}
Air-writing, real-world, benchmark, video-based air-writing Chinese character recognition.
\end{IEEEkeywords}
\section{Introduction}

\label{sec:intro}
\begin{figure}[ht!]
  \centering
  \includegraphics[width=1\linewidth]{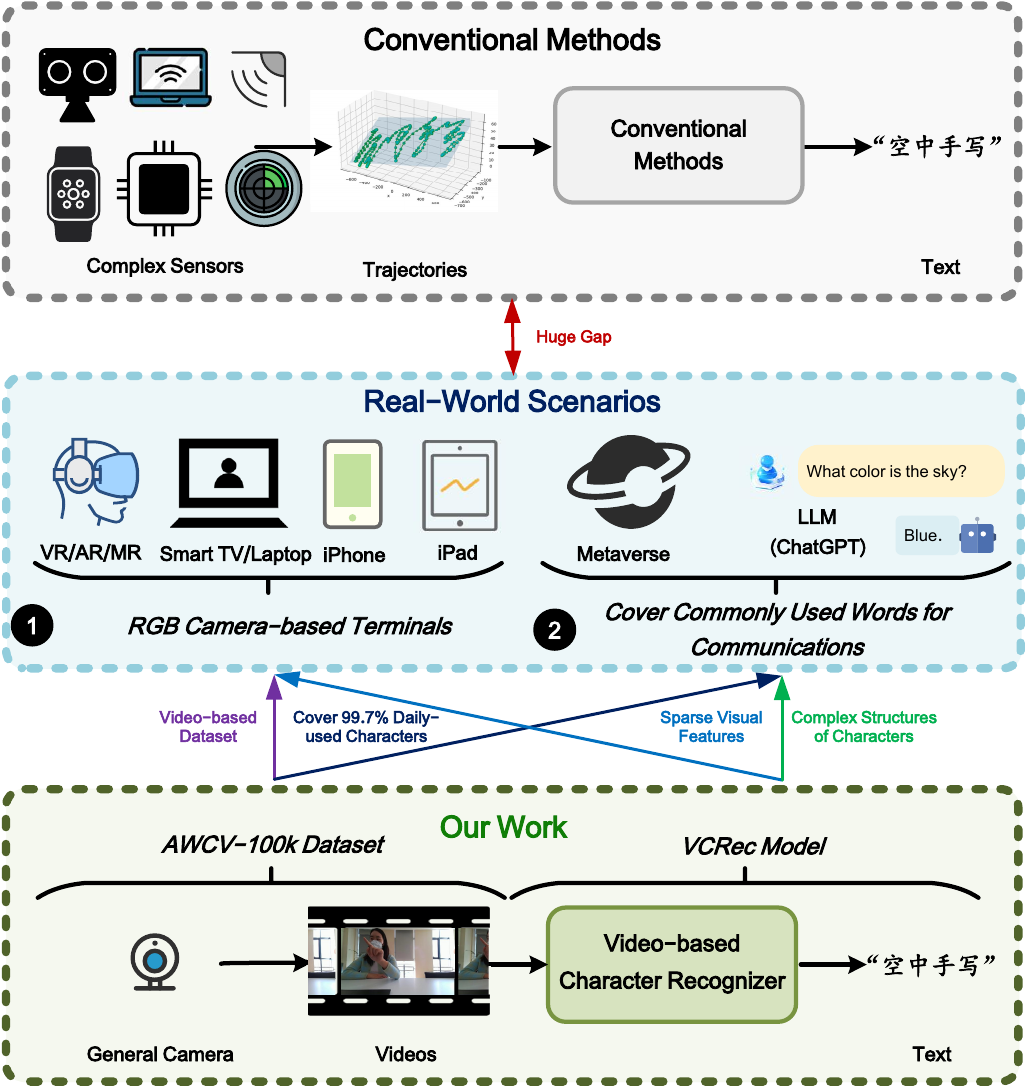}
  \caption{Comparing Our Work and Conventional Air-Writing in Real-World Scenarios. Conventional air-writing relies on accurately captured handwritten trajectories by complex sensors (such as Radar~\cite{9820766}, Smart Watch~\cite{10150241}, Leap Motion~\cite{gan2018unified}, EEGs~\cite{tripathi2023neuroair}, IMU~\cite{zhang2022wearable}), which impose significant limitations for real-world scenarios ($e.g.$, VR/AR/MR, iPhone, metaverse, GPT series~\cite{brown2020language} and others). Mainstream real-world devices only incorporate standard RGB cameras and require coverage of commonly used words for communication purposes. To address these challenges, we propose a video-based air-writing dataset with a comprehensive corpus (covering 99.7\% of daily-used characters), AWCV-100K-UCAS2024, captured by general cameras, and propose a VCRec for sparse visual features and complex character structures.}
  \label{fig:intro}
\end{figure}

\IEEEPARstart{W}{hen} you gesture with your fingertips to write characters in front of the camera, advanced artificial intelligence technology instantly grasps your intentions and initiates the command execution. 
This concept, commonly referred to as \textit{``finger in camera speaks everything''}, epitomizes the essence of air-writing as a powerful technique facilitating effective human-computer interaction. It entails the recognition of characters from handwritten trajectories in 3D space, resembling depictions seen in movies like Ready Player One.

As depicted in Fig.~\ref{fig:intro} (middle), air-writing holds broad application prospects, such as its integration into popular smart devices ( $e.g.$, VR/AR/MR devices, Smart TV, Laptop, iPhone, iPad ), enabling machines to interpret human intentions in a contactless and silent manner. Specifically, air-writing can be used in intelligent conversational systems ($e.g.$, GPT series~\cite{brown2020language}, ChatGLM~\cite{du2021glm}) or the metaverse. Given the diverse nature of real-world scenarios, air-writing systems necessitate the utilization of readily available sensors, such as RGB-based cameras, and the inclusion of frequently used words for effective communication.

However, conventional air-writing methods (Fig.~\ref{fig:intro} upper)~\cite{zhang2013new, chen2015air, huang2016pointing, 6arsalan2021air, xu2015recognition, qu2018data, gan2019air, gan2020compressing} relies on accurately captured handwritten trajectories by complex sensors ($e.g.$, Radar~\cite{9820766}, Smart Watch~\cite{10150241}, Leap Motion~\cite{gan2018unified}, EEGs~\cite{tripathi2023neuroair}, IMU~\cite{zhang2022wearable} ), which are inflexible and challenging to seamlessly integrate into popular smart devices. Moreover, the corpus of some works~\cite{mukherjee2019fingertip, kim2022writing} could not cover commonly used words for communications.

In order to overcome these limitations, we utilize common RGB cameras to record videos of handwritten gestures, enabling practical implementation in real-world scenarios. We create a video-based air-writing benchmark called AWCV-100K-UCAS2024, which consists of a vast collection of 8.8 million video frames. This benchmark encompasses a comprehensive corpus, encompassing 3,755 Chinese characters from the GB1 set, representing 99.7\% of characters commonly used in daily communication~\cite{liu2007word}.

Furthermore, we propose a simple yet effective two-stage solution called the video-based character recognizer (VCRec) (Fig.~\ref{fig:intro} lower) to tackle this challenging task. The key to its success lies in leveraging sparse visual features, which are commonly found in real-world applications due to low frame rates. In the first stage, we introduce a fingertip feature extractor to condense the sparse visual features into fingertip features. In the second stage, VCRec adopts a spatial-temporal sequence module to model the character, capturing temporal information from fingertip movements. Concurrently, we employ a stroke graph attention network (StrokeGAT) to represent the spatial structure of Chinese characters, enhancing the utilization of sparse visual features.

Finally, we include comprehensive quantitative and qualitative evaluations on the AWCV-100K-UCAS2024 benchmark, comparing our approach to existing models in the field of video-based air-writing. Through extensive experimental analysis, we demonstrate that our approach outperforms conventional state-of-the-art (SOTA) methods, achieving a 4.92\% improvement in recognition accuracy on the constructed AWCV-100K-UCAS2024 dataset.

In general, the main contributions of this paper can be summarized as follows:
\begin{itemize}
    \item \textbf{AWCV-100K-UCAS2024 Introduction:} Presented the pioneering video-based air-writing dataset with comprehensive corpus, AWCV-100K-UCAS2024, comprising 8.8 million frames and 3,755 Chinese characters, addressing limitations by utilizing general RGB cameras for real-world applications.
    \item \textbf{VCRec Model Proposition:} Introduced VCRec, a two-stage character recognition model leveraging sparse visual features, achieving improved accuracy in air-writing.
    \item \textbf{Performance Enhancement:} Through extensive analysis, demonstrated a significant 4.92\% accuracy improvement over existing methods on the AWCV-100K-UCAS2024, advancing effective human-computer interaction in real-world.
\end{itemize}

\section{Related Works}
\label{sec:related}

\subsection{Air-Writing Datasets}

Numerous air-writing systems have emerged as a new and intriguing research topic in the field of human-computer interaction, integrating various types of sensors in recent years. Depending on the type of sensor employed, air-writing datasets can be categorized into two main groups: trajectory-based datasets collected by complex sensors and video-based datasets collected by general cameras.
\begin{figure}
\begin{center}
 \includegraphics[width=1\linewidth]{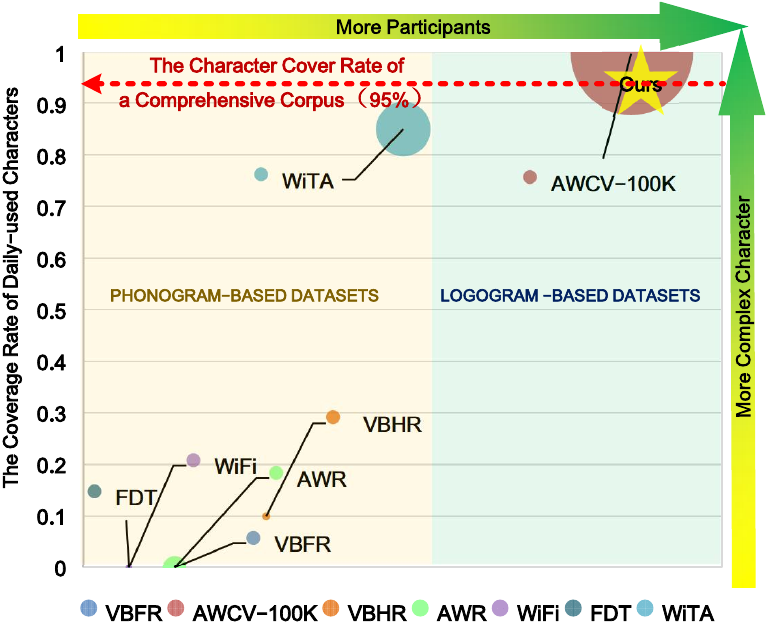}
\end{center}
   \caption{Comparison between AWCV-100K-UCAS2024 with other benchmarks. Phonogram-based (\eg, VBFR~\cite{jin2007novel}, VBHR~\cite{schick2012vision}, AWR~\cite{chen2015air}, WiFi~\cite{fu2018writing}, FDT~\cite{mukherjee2019fingertip}, WiTA~\cite{kim2022writing}) and logogram-based benchmarks are selected for overall comparison. The bubble diameter is proportional to the total frames of the benchmark, and the vertical represents the coverage rate of daily-used characters in each benchmark. Obviously, the proposed AWCV-100K-UCAS2024 is the first logogram-based video dataset with a comprehensive corpus, more participants, and more complex characters.
 }
\label{fig:static}
\end{figure}

\noindent \textbf{Trajectory-based Datasets.}  They were collected by complex sensors ($e.g.$, Radar~\cite{9820766}), data gloves~\cite{fels1993glove, koltringer2007twostick, amma2010airwriting, amma2012airwriting}, hand motion sensors~\cite{amma2014airwriting, chen2015air}, SmartWatch~\cite{10150241}, Wifi~\cite{10.1109/TMC.2018.2831709}, Leap Motion~\cite{gan2018unified}, EEGs\cite{tripathi2023neuroair}, IMU~\cite{zhang2022wearable}, which acquired precise trajectories of air-writing.
 Kumar et al.~\cite{kumar2016study} proposed a 3D English text air-writing system based on Leap Motion, collecting 560 sentences from 10 participants. However, most of them were not available. Qu et al.~\cite{qu2018data} introduced IAHCC-UCAS2016, a trajectory-based air-writing Chinese character dataset. Gan and Wang~\cite{2019In} proposed IAHEW-UCAS2016, a trajectory-based English word air-writing dataset. Gan et al.~\cite{gan2020air} proposed IAHCT-UCAS2018, a trajectory-based air-writing Chinese text dataset. These were public trajectory-based air-writing datasets collected by Leap Motion. With the introduction of public trajectory-based air-writing datasets, air-writing systems based on complex sensors have made some progress. However, due to their high cost and the difficulty of integrating these complex sensors into existing systems, this poses a challenge for their application in real-world scenarios.

\noindent \textbf{Video-based Datasets.} Video-based Datasets were collected by cameras ($e.g.$, Kinect~\cite{zhang2012microsoft}, general RGB camera). Zhang et al.~\cite{zhang2013new} presented a small video-based air-writing dataset that utilized Kinect sensors. Unlike complex and expensive sensors, RGB cameras do not require physical contact, making them a convenient and cost-effective option. As shown in Fig.~\ref{fig:static}, we have summarized video-based air-writing datasets based on participant count, vocabulary categories, and the coverage rate of daily-used characters.
VBFR~\cite{jin2007novel}, VBHR~\cite{schick2012vision}, FDT~\cite{mukherjee2019fingertip} had collected air-writing video datasets for English lowercase letters but have not accessed them. Kim et al.~\cite{kim2022writing} introduced an English and Korean video-based air-writing dataset (WiTA), which was accessed. However, the corpora of the previous datsets did not achieve a coverage rate of over 95\% for `commonly used words', constraining research methodologies for real-world applications. Moreover, there hasn't been a video-based air-writing dataset focusing on logograms in the past, hindering general air-writing system developing.

\begin{figure*}
\begin{center}
  \includegraphics[width=1\textwidth]{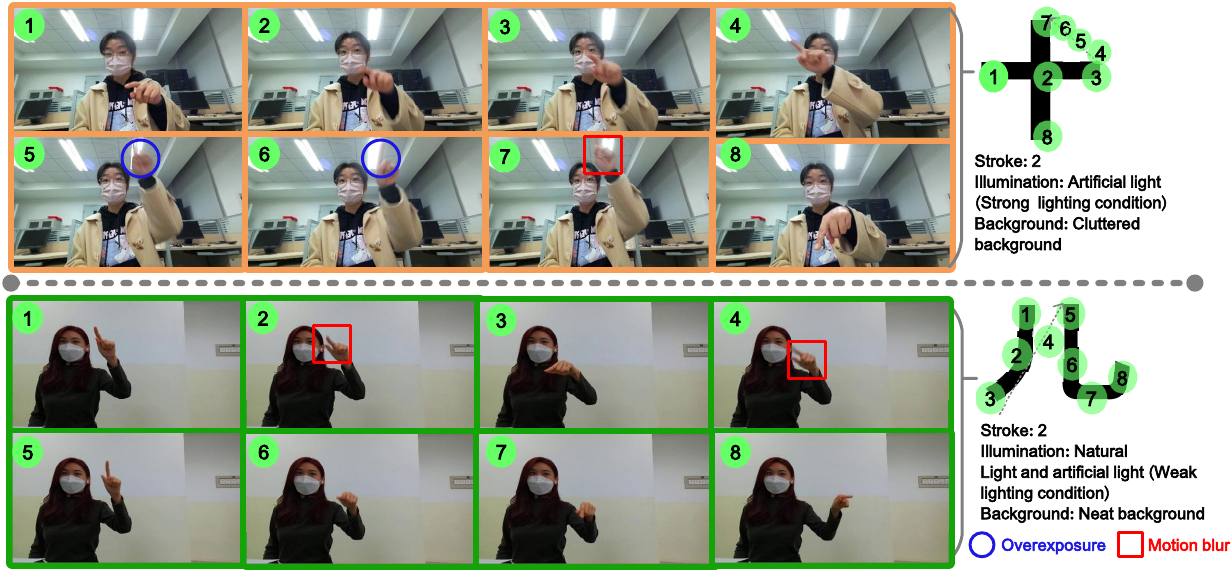}    
\end{center}
\caption{ Examples of AWCV-100K-UCAS2024. The figure shows a comparison of data under different lighting intensities and
backgrounds. On the left of the figure are the video frames of datasets, and on the right are the labels of datasets. The blue circles represent overexposure due to strong illumination and the red box represents motion blur. \textbf{(TOP)} Data is collected under complex backgrounds and strong lighting conditions. \textbf{(BOTTOM)} Data is collected under simpler backgrounds and weaker lighting conditions.}
  \label{fig:dataset-sample}
\end{figure*}

\begin{figure*}[ht!]
  \centering
  \includegraphics[width=1\linewidth]{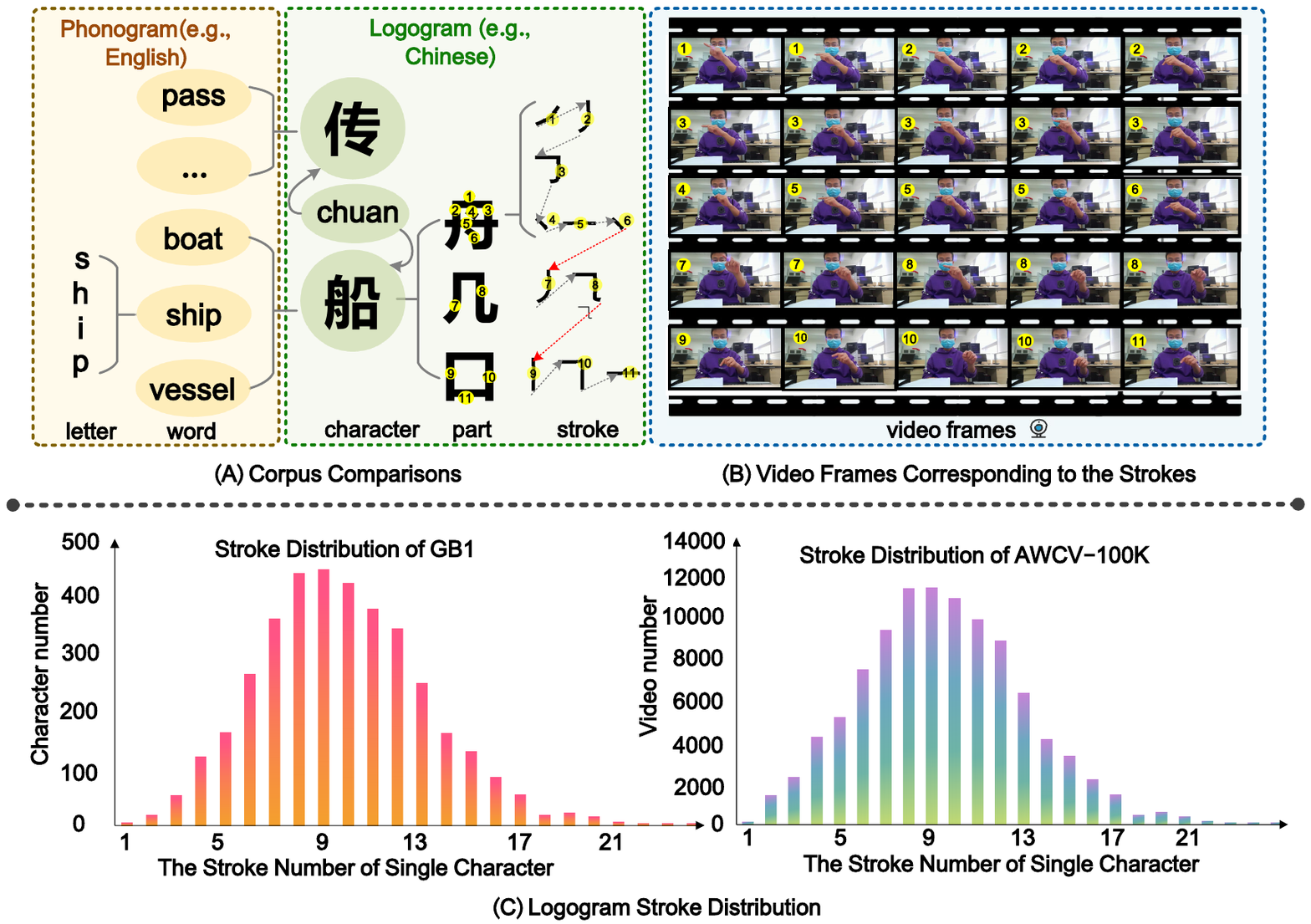}
  \caption{More Complex and Comprehensive Corpus. The figure shows the characteristics of AWCV-100K-UCAS2024. \textbf{(A)} Phonograms are composed of letters and easy to identify (\ie, ``\texttt{ship}" consists of the letters ``\texttt{s,h,i,p}" ). Logograms are described as having one pinyin corresponds to multiple characters. Each character is made up of parts, which can be further divided into strokes (\ie, pinyin ``\texttt{chuan}" with yellow numbers representing the stroke order). \textbf{(B)} Successive frames represent a stroke (\ie, the first two frames of the video correspond to the first stroke in figure), which are collected by general camera under cluttered background and natural light. \textbf{(C)}The stroke distributions of GB1 and AWCV-100K-UCAS2024 respectively.}
  \label{fig:dataset}
\end{figure*}

\subsection{Air-Writing Recognition Models}

Most conventional air-writing recognition methods heavily rely on precise handwritten trajectories. For instance, Zhang et al.~\cite{zhang2013new} utilized Kinect sensors for fingertip tracking, enabling controller-free motion tracking. Their system primarily focused on recognizing characters using the modified quadratic discriminant function (MQDF) classifier. In another approach, Kumar et al.~\cite{kumar2016study} segmented each text into individual words and employed an LSTM-CTC structure for word recognition. Similarly, Gan and Wang~\cite{gan2019air} applied an LSTM-based sequence-to-sequence model for word recognition, achieving performance comparable to previous state-of-the-art methods. Gan et al.~\cite{GAN2023109317} revolutionized character representation by adopting skeleton graphs and introducing PyGT, a specialized transformer-convolutional network fusion. Furthermore, Wu et al.~\cite{wu2023air} introduced the attention convolutional loop network (ACRN), utilizing 1DCNN feature extraction followed by LSTM multi-head attention mechanism classification. Their experiments on CASIA-OLHWDB2.0-2.2~\cite{6065272} and IAHCT-UCAS2018~\cite{gan2020air} demonstrated higher recognition accuracy. 

Recently, video-based air-writing was proposed by Kim
et al.~\cite{kim2022writing}, introducing residual network architectures inspired by 3D ResNet. Additionally, Tan et al.~\cite{tan2023end} proposed transformer architectures for air-writing recognition. However, these methods did not particularly focus on the visual semantics and spatial features of fingertips, which led to lower performance.


\subsection{Fingertip Detection and Tracking}
In the realm of human-computer interaction (HCI), fingertip detection and tracking have been explored. Initially, Liang
et al.~\cite{liang20123d} employed palm-to-hand outline distance measures for fingertip identification, further refined~\cite{krejov2013multi} using the hand's natural structure. However, challenges persisted with segmentation quality due to reliance on cues like color, depth, and motion~\cite{zhang2013new}. Preceding these methods were model-based 3D gesture tracking approaches~\cite{de2011model, chang2016spatio}, though these demanded significant computational resources and ample training data, limiting their real-time applicability. Contrastingly, MediaPipe, an open-source gesture recognition framework, offered real-time detection of 21 key point coordinates across various platforms, achieving impressive frame rates like 909 FPS on the iPhone 11, with a mean square error of 9.817mm~\cite{lugaresi2019mediapipe, zhang2020mediapipe}. Recently, in the field of visual object tracking, researchers have devised various strategies such as dynamic attention-guided multi-trajectory methods~\cite{wang2021dynamic}, dynamic feature assignment frameworks like DFAT~\cite{DBLP:journals/tcsv/ZhangJLYL23}, occlusion-aware networks like SiamON~\cite{DBLP:journals/tcsv/FanYHSWL23}, and the Siamese-based Twin Attention Network~\cite{DBLP:journals/tcsv/BaoSZL23}, alongside simplified long-term tracking~\cite{DBLP:journals/tcsv/XuZWS23}, which could bolster gesture recognition and improve object manipulation by complementing existing fingertip detection and tracking methods. 


\section{AWCV-100K-UCAS2024 Dataset}
\label{sec:AWCV-100K-UCAS2024}

\begin{figure*}
\begin{center}
 \includegraphics[width=0.9\linewidth]{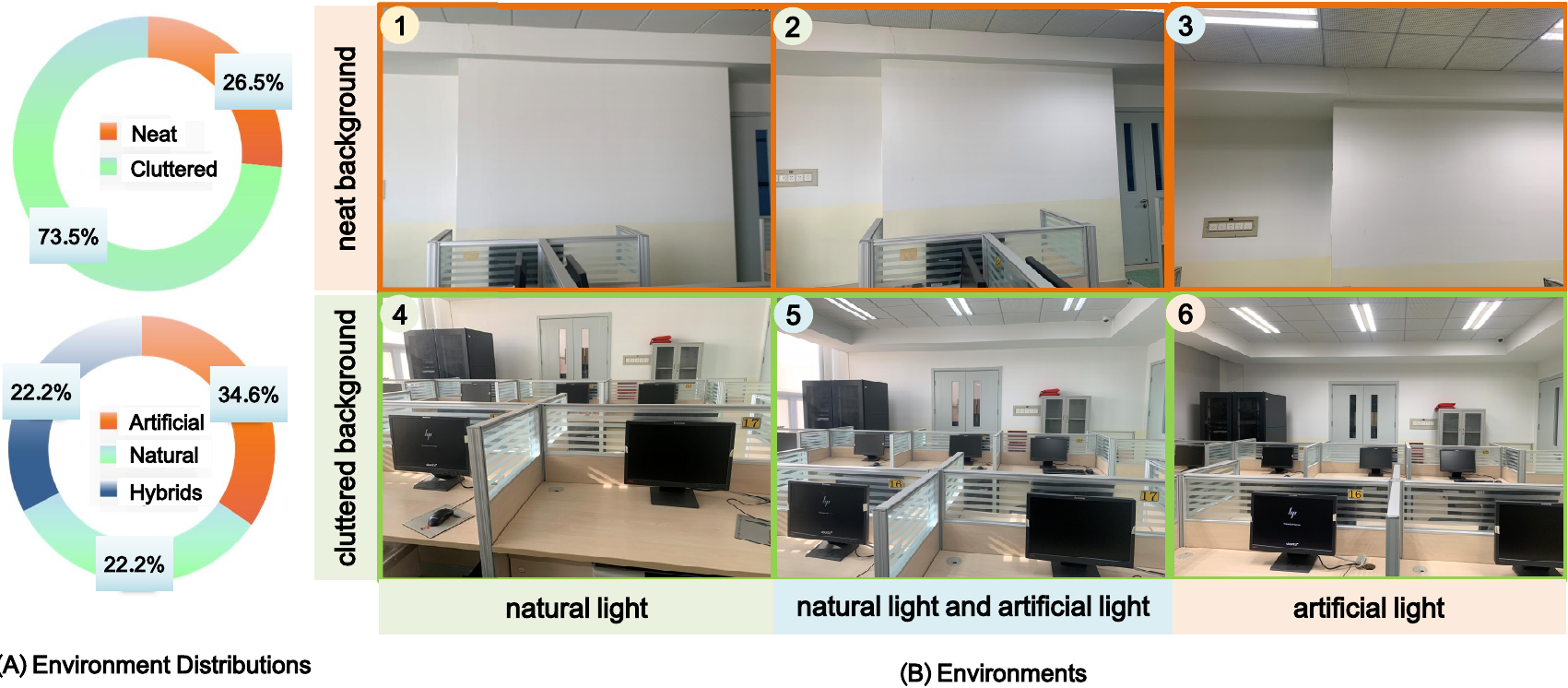}
\end{center}
   \caption{
Statistical Analysis Environments in AWCV-100K-UCAS2024. (A) Depicts the diverse environmental distributions within AWCV-100K-UCAS2024. (B) This section illustrates the environmental combinations of two backgrounds (\ie, neat background and cluttered background) and three types of light (\ie, natural light, artificial light, and their hybrids).
}
\label{fig:environment and data statistic}
\end{figure*}

Mainstream real-world devices typically utilize standard RGB cameras and necessitate comprehensive coverage of commonly used words for effective communication. To tackle these challenges, we introduce AWCV-100K-UCAS2024, a pioneering video-based air-writing dataset designed for real-world applications. This extensive dataset is collected using general cameras and involves a large number of participants.

\subsection{Data Collection}
We have utilized an air-writing platform for data acquisition. Initially, we have briefed the participants on the data collection procedure. They have been instructed to assume that a perfect AI system would decode their air-writing, encouraging them to write as naturally as possible. Each participant has then composed approximately 500 words in Chinese, resulting in a total collection of 102,688 videos. During each data collection session, participants have had the flexibility to adjust the camera view, accommodating different angles and positions. The image sequences have been derived from real-time recorded videos at 30 frames per second (FPS). The examples of AWCV-100K-UCAS2024 are shown in Fig.~\ref{fig:dataset-sample}.

\subsection{Checkout Flow}

We have implemented a stringent data review process to uphold the benchmark's quality. Trained professional collectors understand the nuances of the air-writing task and undertake preliminary work with a self-inspection process. Subsequently, verifiers conduct a second-round review of the collected data. Finally, authors make the ultimate judgment whether to accept or reject the data in the third-round confirmation. Any rejection during self-check, verification, or data acceptance necessitates recollection. We believe this three-round verification mechanism ensures the generation of a high-quality dataset.

\subsection{Challenge Attributes}

\noindent\textbf{Sparse Visual Features.} 
In real-world scenarios, mainstream devices only incorporate standard
RGB cameras. In order to be more widely applicable in the real-world, we collect datasets by general RGB cameras, as illustrated in Fig.~\ref{fig:dataset} (B). However, compared to sophisticated sensors (such as Leap Motion, which operates at 120FPS), general cameras often capture fewer frames per second ($e.g.$, 30FPS), resulting in the acquisition of relatively sparse features.

\begin{table}[t!]
\centering
\begin{small}
\caption{Summary of the participant information statistics. Statistics are gathered on participants' age, gender, handedness, and hand size.}
\label{tb:user_statistics}
\begin{tabular}{llll}
\toprule

\bf{Metric} & \bf{Type} & \bf{Value} \\
\midrule
\multirow{3}{*}{{\bf{Gender}}} & Male & $145$/$211$\\
& Female & $66$/$211$ \\
& Neutral & $0$/$211$ \\
\midrule
\multirow{3}{*}{{\bf{Age}}} & Range & $20$ - $30$ \\
& Average & $23.33$\\
& s.t.d. & $1.39$ \\
\midrule
\multirow{3}{*}{{\bf{Comfort-Hand}}} & Left & $1$/$211$\\
& Right & $210$/$211$ \\
& Both & $0$/$211$ \\\midrule
\multirow{3}{*}{{\bf{Hand Width}}} & [6cm, 8cm] & $67$/$211$\\
& (8cm, 10cm] & $140$/$211$ \\
& $>$ 10cm & $24$/$211$ \\\midrule
\multirow{4}{*}{{\bf{Hand Length}}} & [15cm, 17cm] & $15$/$211$\\
& (17cm, 19cm] & $60$/$211$ \\
& (19cm, 20cm] & $131$/$211$ \\
& $>$ 20cm & $5$/$211$ \\
\bottomrule
\end{tabular}
\end{small}
\end{table}

\begin{table*}[ht!]
\centering
\begin{small}
\caption{Comparison of datasets. The proposed AWCV-100K-UCAS2024 dataset is the most comprehensive and provides rich types of data instances. Our dataset supplies videos containing semantic text written in the air, which captures the interdependence between gestures for different characters. ( CRD, Sem, K, E, C, and N in the table stand for inclusion of coverage rate of daily-used characters, semantic words, Korean, English, Chinese, and numbers, respectively.)}
\label{tb:dataset_comparison}
\begin{tabular}{lllllllllll}
\toprule
\multicolumn{1}{l}{\textbf{Dataset}} & \textbf{Year} & \multicolumn{1}{l}{\textbf{People}}  & \textbf{Frames} & \textbf{CRD} & \textbf{Sem} & \textbf{Language}  & \textbf{Sensor} & \textbf{Illumination} & \textbf{Access}\\
\midrule
VBFR \cite{jin2007novel}           & 2007 & 69  & -      & -  & -  & E        & RGB    & -         & -\\
VBHR \cite{schick2012vision}       & 2012 & 21  & -      & -  & -  & E        & RGB    & -         & -\\
ANWE \cite{zhang2013new}           & 2013 & -   & 44,522 & -  & -  & ECN      & Depth  & -              & -\\
AWR \cite{chen2015air}             & 2015 & 22  & -      & -  & \checkmark  & E  & Motion  & -         & -\\
PGEI \cite{huang2016pointing}      & 2016 & 24  & 93,729 & -  & -  & EC       & Depth & - & -\\
WiFi \cite{fu2018writing}          & 2018 & 5   & -      & -  & -  & E       & WiFi   & -         & -\\
FDT \cite{mukherjee2019fingertip}  & 2019 & 5   & -      & -  & -  & EN        & RGB    & -              & -\\
WiTA \cite{kim2022writing}         & 2021 & 122 & 1,757,307 & 89.9\%  & \checkmark & EK  & RGB     & - & \checkmark\\
\midrule
\textbf{AWCV-100K-UCAS2024 (Ours)}          & 2023 &  \textbf{211}   &   \textbf{8,819,068}  & \textbf{99.7\%}  & \checkmark & C & RGB     &  \checkmark & \checkmark\\
\bottomrule
\end{tabular}
\end{small}
\end{table*}

\noindent\textbf{More complex corpus.} 
Phonograms usually consist of only a few dozen letters ($e.g.$, English has 26 letters, German has 27, Russian has 33, \etc). Logograms, on the other hand, comprise thousands of characters ($e.g.$, Chinese has 3,755 characters only in the GB1 set, \etc). However, previous video-based air-writing datasets focused on phonograms and did not include a comprehensive corpus (a corpus vocabulary size achieving a coverage rate of over 95\% for `commonly used words'). To bridge this gap, we have constructed AWCV-100K-UCAS2024 with a comprehensive logogram corpus. It includes 3,755 Chinese characters from the GB1 set, encompassing 99.7\% of characters used in daily communication~\cite{liu2007word}, forming a comprehensive corpus. As shown in Fig.~\ref{fig:dataset} (A), Chinese characters have complex structures, which are composed of multiple parts, each consisting of many strokes. The stroke distributions of GB1 and AWCV-100K-UCAS2024 are shown in Fig.~\ref{fig:dataset} (C), which refers the stroke number of character is various. Moreover, Chinese characters are formed in various ways, resulting in highly intricate structures. These factors have presented greater challenges for video-based air-writing recognition.

\noindent\textbf{More Various Environments.}
As shown in Fig.~\ref{fig:environment and data statistic}, we have collected data in various environments, encompassing different lighting conditions ($e.g.$, artificial light, natural light, and a combination of both) and backgrounds ($e.g.$, a neat background and a cluttered background) to ensure robustness against real-world scenario variations. Additionally, during data collection, we specifically have focused on capturing data under diverse weather conditions and at different times of the day (morning, afternoon, evening) to more accurately simulate natural light variations in real-world environments. Furthermore, participants have been allowed to adjust their seats, cameras, \textit{etc.} We have varied viewpoints (camera distance, angle, and position) during different data collection processes to enhance the diversity of AWCV-100K-UCAS2024.

\noindent\textbf{More Participants.} 
As depicted in Table~\ref{tb:user_statistics}, which summarizes participant statistics, we have recruited a total of 211 individuals (male: 145, female: 66), who are native Chinese speakers proficient in both reading and writing. Participants' hand lengths have ranged from 15.7 cm to 22.5 cm (M=18.29 cm, SD=1.05 cm), and their hand widths have ranged from 6.1 cm to 10.5 cm (M=8.16 cm, SD=0.73 cm). The diverse hand sizes and writing styles among numerous participants pose challenges to the accuracy of video-based air-writing recognition.

\noindent\textbf{Others.} 
As depicted in Fig.~\ref{fig:dataset-sample}, rapid fingertip movements resulting in motion blur and problems arising from excessive illumination create hurdles in acquiring precise trajectory features. These issues specifically affect the image's sharpness and contours, adding complexity to the recognition process.

\subsection{Dataset Comparison.} 

Table~\ref{tb:dataset_comparison} presents a summary of the air-writing datasets collected in this study, comparing them with previous studies and highlighting the significant advantages of our dataset. 
\textit{\textbf{(1) Originality:}} Our dataset stands as the sole publicly accessible study focusing on logograms, offering invaluable support for research into video-based air-writing in real-world scenarios. 
\textit{\textbf{(2) Comprehensiveness:}} Our dataset achieves comprehensive coverage of the Chinese corpus by encompassing all characters of the GB1 set. In contrast, other datasets primarily concentrate on phonograms and provide limited coverage via select word videos, posing challenges in real-world applications. 
\textit{\textbf{(3) Authenticity and Diversity:}} Our dataset spans a wide array of acquisition dimensions necessary for real-world applications, including scenes captured with general cameras. Additionally, we encompass diverse environments and various illuminations to closely mimic realistic scenarios. Moreover, by incorporating participants with diverse hand sizes and writing styles, our dataset mirrors the diversity among users. These elements present significant challenges while providing researchers with invaluable insights to study algorithm robustness in real-world scenarios.

\subsection{Evaluation Protocol}

To measure the recognition performance, both the correct rate (CR) and the accurate rate (AR) are used as the performance metrics, defined in the ICDAR 2013 Chinese hand-writing recognition competition~\cite{yin2013icdar}. Specifically,
\begin{equation}
    C R=\left(N_{t}-D_{e}-S_{e}\right) / N_{t},
\end{equation}
\begin{equation}
    A R=\left(N_{t}-D_{e}-S_{e}-I_{e}\right) / N_{t},
\end{equation}
where $N_{t}$ is the total number of characters in the test ground-truth sentences, while $D_{e}$, $S_{e}$, $I_{e}$ denote deletion error, substitution error, and insertion error, respectively, between predictions and test ground-truth sentences. Additionally, in this paper, the sequence length of characters is set to 1.

\begin{figure*}
\begin{center}
  \includegraphics[width=1\textwidth]{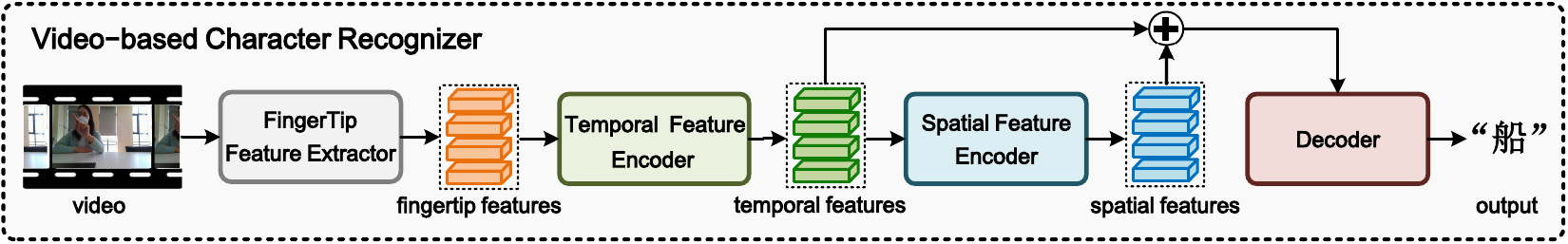}    
\end{center}
\caption{Overview of the Video-based Character Recognizer (VCRec). The VCRec comprises a Fingertip Feature Extractor, a Spatio-Temporal Sequence Module that encompasses both Temporal Feature Encoder and Spatial Feature Encoder, as well as a CTC Decoder.}
  \label{fig:framework}
\end{figure*}


\section{Methodology}

\subsection{Overview}

As shown in Fig.~\ref{fig:framework}, we propose the Video-based Character Recognizer (VCRec), a two-stage method that is simple yet effective in addressing sparse visual features. The video initially undergoes adaptive fingertip feature extraction via the fingertip feature extractor (Sec.~\ref{sec:ffe}). Subsequently, these adaptive fingertip features are processed within the spatio-temporal sequence module (Sec.~\ref{sec:stsm}), where they are encoded to extract intrinsic characteristics along two distinctive dimensions. 
More specifically, the temporal feature encoder captures temporal aspects of the fingertip, while the spatial feature encoder handles the fingertip's spatial features. Finally, the decoder decodes the fusion features into the character.

\subsection{Fingertip Feature Extractor} 
\label{sec:ffe}
\begin{figure}
\begin{center}
 \includegraphics[width=1\linewidth]{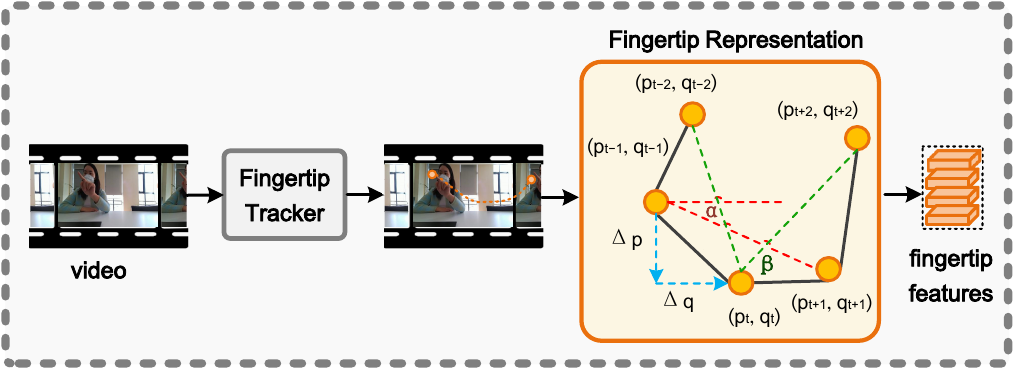}
\end{center}
   \caption{ Fingertip Feature Extractor. The video is first inputted by the Fingertip Tracker to obtain fingertip trajectories and encoded into fingertip features by Fingertip Representation.
   }
\label{fig:ffextractor}
\end{figure}

Due to the low frame rate in real-world scenarios, the key lies in utilizing sparse visual features. To address this challenge, we introduce a Fingertip Feature Extractor to compress sparse visual features into fingertip features. As shown in Fig.~\ref{fig:ffextractor}, the video is first inputted by the Fingertip Tracker~\cite{lugaresi2019mediapipe, zhang2020mediapipe} to obtain fingertip trajectories and encoded into fingertip features by Fingertip Representation.

We represent each trajectory with its derivatives including the offsets of positions and the writing directions rather than its raw absolute coordinates, which can effectively describe the next movement of strokes. As shown in Fig.~\ref{fig:ffextractor}, the following representations are calculated for the {$t$}-th point {$(p_{t},q_{t},s_{t})$} of the trajectory: (1) the offsets of XY-coordinates, {$\Delta p$} and {$\Delta q$}; (2) the cosine and sine of the writing direction {$\alpha$}; (3) the cosine and sine of the curvature {$\beta$}; (4) the change of the stroke identity. As a result, each point {$(p_{t},q_{t},s_{t})$} is represented as an eight-dimensional vector {$\mathbf{x}_{t}$} at time step $t, t\in\mathbb{Z}$, $i.e.$,

\begin{equation}
\begin{aligned}
         \mathbf{x}_{t}=\left[\Delta p_{t}, \Delta q_{t}, \sin \alpha, \cos \alpha, \sin \beta, \cos \beta, \right. \\ 
         \left.\mathbb{I}\left(s_{t}=s_{t+1}\right), \mathbb{I}\left(s_{t} \neq s_{t+1}\right)\right].
\end{aligned}
\end{equation}

\subsection{Spatio-Temporal Sequence Module}
\label{sec:stsm}
The Spatio-Temporal Sequence Module encodes the fingertip features of the character into probabilities as:
\begin{align}
    \textbf{p} & = f_{\text{c}}(f_{\text{g}}(f_{\text{r}}(\textbf{x})),f_{\text{r}}(\textbf{x})),
\end{align}
where $\textbf{x}$ is the fingertip features predicted by the Fingertip Feature Extractor, $f_r$ is the Temporal Feature Encoder, $f_g$ is the Spatial Feature Encoder, and $f_c$ is a decoder head that maps the deep feature into the character probabilities $\textbf{p}$. During training, the cross-entropy loss supervises $\textbf{p}$.
During inference, the character with the highest probability is selected as the prediction.

\begin{figure}
\begin{center}
 \includegraphics[width=1\linewidth]{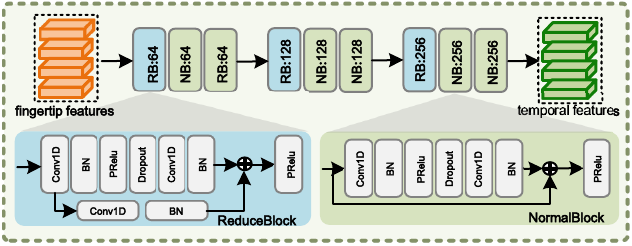}
\end{center}
   \caption{Temporal Feature Encoder. It is constructed as a hierarchical framework, which is primarily composed of ReduceBlock and NormalBlock. 
   }
\label{fig:temporal}
\end{figure}

\noindent\textbf{Temporal Feature Encoder} is uniquely tailored for encoding the temporal information of the fingertip feature sequences, utilizing mainly 1D convolution operators. As shown in Fig.~\ref{fig:temporal}, Temporal Feature Encoder is constructed as a hierarchical framework, which is primarily composed of ReduceBlock and NormalBlock inspired by Gan et al.~\cite{gan2019air}. 
Both types of conv-blocks utilize the techniques, like the residual connection~\cite{he2016deep}, batch normalization~\cite{pmlr-v37-ioffe15}, dropout~\cite{srivastava2014dropout}, and parametric rectified linear unit (PReLU)~\cite{he2015delving}, to speed up the network training and also address the over-fitting problem. The ReduceBlock, contains an extra convolution branch to adopt the residual connection when the numbers of input and output channels are different; if the convolution stride is set to 2, the ReduceBlock will downsample the sequence over the time dimension to increase the receptive field of the convolution.

Through Temporal Feature Encoder, the fingertip feature sequence is encoded by $\textbf{Z}= f_r(\textbf{x})$, $\textbf{Z}=[\textbf{z}_1,\textbf{z}_2,\cdots,\textbf{z}_l]^{\text{T}},\textbf{Z}\in \mathbb{R}^{l\times c}$, where $l$ represents the length of feature in the time dimension and $\{\textbf{z}_i\}_{i\in 1,2\cdots,l}$ are the features of different clips of the fingertip trajectory.

\begin{figure}
\begin{center}
 \includegraphics[width=1\linewidth]{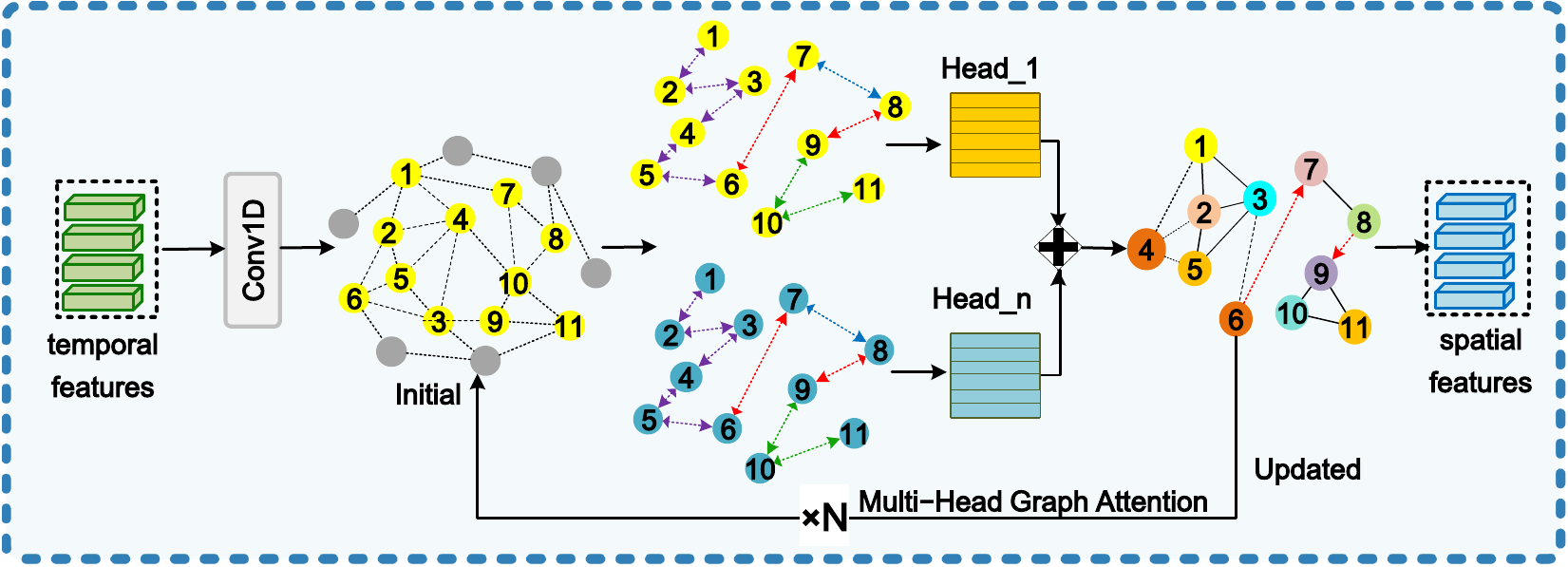}
\end{center}
   \caption{
 Spatial Feature Encoder. It models the spatial structure of the character through a graph attention network, we have named it \textbf{StrokeGAT}.
   }
\label{fig:spatial}
\end{figure}

\noindent\textbf{Spatial Feature Encoder} is used to encode the spatial information of the fingertip feature sequences based on feature $\textbf{Z}$.
As shown in Fig.~\ref{fig:spatial}, by treating the features of different clips of the trajectory ($\ie$, $\{\textbf{z}_i\}_{i\in 1,2\cdots,l}$) as the graph nodes, the Spatial feature encoder models the spatial structure of the character through a graph attention network~\cite{velickovic2017graph}, we have named
it \textbf{StrokeGAT}. 
Through StrokeGAT, the temporal feature $\textbf{Z}$ is encoded by $\overline{\textbf{Z}}=f_g(\textbf{Z})$, where $\overline{\textbf{Z}}=[\overline{\textbf{z}}_1,\overline{\textbf{z}}_2,\cdots,\overline{\textbf{z}}_l]^{\text{T}},\overline{\textbf{Z}}\in \mathbb{R}^{l\times c}$.

Specifically, the decoder head is used to add the temporal features and spatial features of the fingertip movement trajectory and map them into probabilities of characters.
To achieve a faster inference speed, we simply add the two types of features and then utilize a single-layer fully connected layer to map the feature to $\textbf{p}$, $\ie$,  $\textbf{p}=f_c(\overline{\textbf{Z}}\oplus\textbf{Z})=\text{FC}(\frac{1}{l}\sum_{i=1}^l [\overline{\textbf{z}}_i^{\text{T}}\oplus\textbf{z}_i^{\text{T}}])$, where $\text{FC}$ and $[ \oplus ]$ denote the fully connected layer and add operator respectively.

\section{Experiments}

\subsection{Datasets}
Our proposed baseline method VCRec underwent comparisons across the following datasets: the video-based air-writing datasets AWCV-100K-UCAS2024 (Ours), WiTA~\cite{kim2022writing}, and the trajectory-based air-writing datasets IAHCC-UCAS2016~\cite{xu2015recognition} and IAHEW-UCAS2016~\cite{2019In}.

\noindent\textbf{AWCV-100K-UCAS2024} is a video-based air-writing Chinese character dataset, including 102,688 videos and comprising a total of 8.8 million video frames. The benchmark includes 3,755 Chinese characters from the GB1 set, encompassing 99.7\% of characters used in daily communication, forming a comprehensive corpus. To ensure the developed model is robust to variations among different individuals, we have partitioned the dataset into three sets (\ie, training, validation, and testing) in an approximate ratio of 8:1:1, dividing the data by person.

\noindent\textbf{WiTA}~\cite{kim2022writing} is a video-based air-writing photogram dataset. Only the English portion was used in this experiment, which included 10,620 video sequences from 122 participants. The data were sourced from an RGB camera with a frame rate of 29 fps, and all video frames were converted to $224\times224$ pixel images. 

\noindent\textbf{IAHCC-UCAS2016}~\cite{xu2015recognition} is a public trajectory-based air-writing Chinese
character dataset, where each character is written in the midair within a single stroke. The dataset contains 431,825 samples of 3,755 different Chinese characters.

\noindent\textbf{IAHEW-UCAS2016}~\cite{2019In} is a public large-vocabulary trajectory-based air-writing English word dataset. The dataset is contributed by 324 different participants and contains 150,480 recordings covering 2,280 English words.

\begin{figure}
\begin{center}
 \includegraphics[width=1\linewidth]{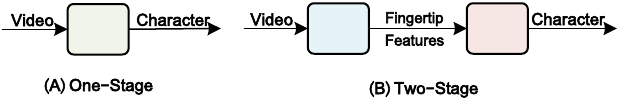}
\end{center}
   \caption{Architectures of Air-Writing Recognition Method. (A) The one-stage method refers to encoding the features of the video and then decoding the characters. (B) The two-stage method refers to first extracting the fingertip trajectories, and then encoding the fingertip trajectories. Finally, decode the characters.
   }
\label{fig:stage}
\end{figure}

\subsection{Implementation Details} 
The experiment utilizes a Spatio-Temporal Sequence Module with a dropout probability of 0.2 in each conv-block and fully connected layer to ensure generalization. As shown in Fig.~\ref{fig:spatial}, $N$ is 1, that is, one graph attention layer is used. $n$ is 8, that is, using the 8-head self-attention mechanism. The proposed architecture is implemented using PyTorch~\cite{paszke2017pytorch} and initialized with default parameters after resizing images to $112\times112$. Optimization is achieved using the Adam~\cite{kingma2014adam} algorithm with a mini-batch size of 8. The initial learning rate is set to 0.001 and is decreased by a factor of 0.01 when recognition performance plateaus. The experiments are conducted on 4 NVIDIA TITAN RTX 24G GPUs.

\subsection{Results and Analysis}

\begin{table}[]
\centering
\caption{Results of Different Methods on AWCV-100K-UCAS2024.}
\label{table:SOAT}
\resizebox{\linewidth}{!}
{
\begin{tabular}{llllll}
\toprule
\multicolumn{2}{c}{\textbf{Method}}           &  \textbf{Architecture}     &\textbf{AR(\%) $\uparrow$} & \textbf{Params(M) $\downarrow$} & \textbf{FPS $\uparrow$} \\ \midrule
                                     & \textbf{CNN+LSTM}~\cite{2017TheKinetics}  &       & 3.31            & 43.2  &  20.2 \\
                                     & \textbf{TwoStream}~\cite{2017TheKinetics} &       & 5.64            & 62.6  &  48.7 \\
                                     & \textbf{C3D}~\cite{2017TheKinetics}       &       & 4.71            & 93.3  & 30.3  \\
                                     & \textbf{ST-MC}~\cite{kim2022writing}     &        & 14.25            & 17.5 &  272.9 \\
                                     & \textbf{ST-rMC}~\cite{kim2022writing}     &       & 16.02            & 52.4 &   306.7\\
                                     & \textbf{ST-R(2+1)D}~\cite{kim2022writing} &       & 4.51             & 52.4 &   126.3\\
                                     & \textbf{ST-R3D}~\cite{kim2022writing}     & \multirow{-5}{*}{\textbf{CNN/RNN}}     & 23.40 & 55.4  & 168.7\\\cmidrule(l){2-6}
\multirow{-7}{*}{\textbf{One Stage}}  & \textbf{ViT}~\cite{dosovitskiy2020image}  &   \textbf{Transformer}  & 21.51           & 86.7 &  16.9\\\midrule

\textbf{Two Stage}                 & \textbf{VCRec (Ours) }                    & \textbf{CNN+GAT}                       & \textbf{52.43}    & 3.9  & 88.2 \\\bottomrule
\end{tabular}
}
\end{table}

\begin{figure*}
\begin{center}
 \includegraphics[width=0.9\linewidth]{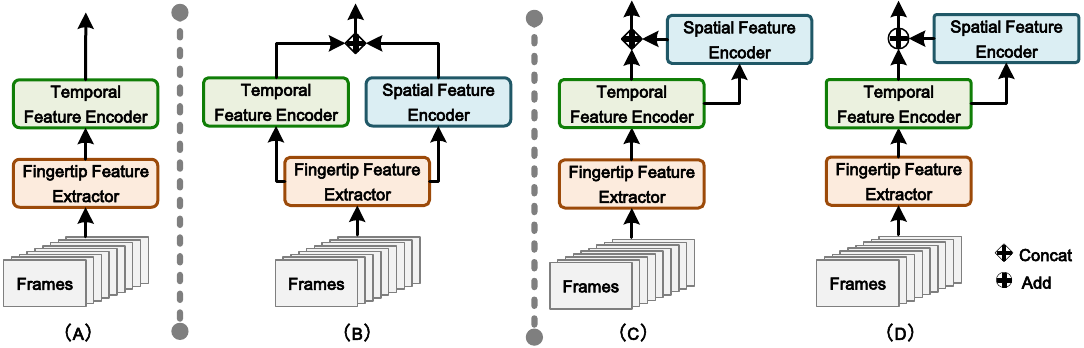}
\end{center}
   \caption{
   Feature Fusion Strategy. (B), (C) and (D) are three different architectures of feature fusion strategies. (A) is the model without a Spatial Feature Encoder.
   }
\label{fig:featurefusionmode}
\end{figure*}

\noindent\textbf{Comparison of Different Architectures.}
As shown in Fig.~\ref{fig:stage}, methods are categorized into two groups based on their utilization of fingertip features: one-stage (not using fingertip features) and two-stage (using fingertip features).
For video-based air-writing recognition tasks, one-stage methods ($e.g.$, ST-R3D, ST-(2+1)D~\cite{kim2022writing}) encode the video frames directly for character recognition. 
To address sparse visual features, we first propose the two-stage architecture, which refers to extracting the fingertip features from sparse visual features and then encoding the fingertip features for character recognition. 

Table~\ref{table:SOAT} illustrates the performance of different model architectures on the AWCV-100K-UCAS2024 dataset. For the one-stage architecture, aside from ST-R3D~\cite{kim2022writing}, we have conducted experiments with various classic architectures ($e.g.$, C3D~\cite{2017TheKinetics}, TwoStream~\cite{2017TheKinetics} and so on) used for modeling video sequences. Additionally, we have also utilized the ViT architecture. Several insights can be derived from these findings. Our proposed two-stage method significantly outperforms the one-stage approach, highlighting the crucial importance of fingertip features in air-writing recognition.
Compared with ST-R3D~\cite{tan2023end}, the previous SOTA method in video-based air-writing, VCRec (Ours) enhances accuracy by 29.03\% on the AWCV-100K-UCAS2024 dataset. ViT architecture has been not very effective, possibly because the transformer architecture emphasizes global features but struggles to focus on specific visual characteristics. In the case of air-writing, effectively capturing visual features is particularly challenging due to their sparsity, thus leading to difficulty in achieving optimal performance.

\begin{table}[]
\centering
\caption{Results of Different ST Methods on AWCV-100K-UCAS2024. S represents the Spatial Feature Encoder and T represents the Temporal Feature Encoder.}
\label{table:st}
\begin{small}
\begin{tabular}{lcll}
\toprule
\textbf{Method}                    & \textbf{ST}  &\textbf{AR(\%) $\uparrow$} & \textbf{Params(M) $\downarrow$}              \\ \midrule
\textbf{1D-TCRN}~\cite{gan2020air} & \textbf{T}  &45.31  & 5.6 \\
\textbf{LSTM}~\cite{LSTM}          & \textbf{T}  & 43.23  &6.5 \\
\textbf{1DCNN}~\cite{gan2019new}   & \textbf{T}  & 47.51 & 1.1 \\
\textbf{Transformer}~\cite{DBLP:journals/corr/VaswaniSPUJGKP17}               & \textbf{T}  & 40.11  &  36.2\\ 
\midrule
\textbf{VCRec (Ours)}              & \textbf{ST} &\textbf{52.43} & 3.9 \\ \bottomrule
\end{tabular}
\end{small}
\end{table}

\begin{table}
\centering
\caption{Results of Different Feature Fusion Strategies on AWCV-100K-UCAS2024. A, B, C, and D correspond to the four different structures in Fig.~\ref{fig:featurefusionmode}.}
\label{table:FeatureFusionMode}
\begin{small}
\begin{tabular}{lll}
\toprule
\multicolumn{1}{l}{\textbf{Feature Fusion Strategy}} &\textbf{AR(\%) $\uparrow$} & \textbf{Params(M) $\downarrow$} \\ \midrule
\textbf{A} &47.51  & 1.1 \\ 
\textbf{B}                & 48.19 & 3.7 \\ 
\textbf{C}                & 51.77  &3.9 \\ \midrule
\textbf{D(Ours)}          & \textbf{52.43} & 3.9 \\ \bottomrule
\end{tabular}
\end{small}
\end{table}

\noindent\textbf{Comparison of Different Spatio-Temporal Sequence Modules.}
As shown in Fig.~\ref{fig:featurefusionmode}, we design different Spatio-Temporal Sequence Module architectures. Fig.~\ref{fig:featurefusionmode} (A) is the first structure without a spatial feature encoder. Fig.~\ref{fig:featurefusionmode} (D) is the structure of VCRec (Ours). We conduct some experiments shown in Table~\ref{table:st}. The Transformer model we're using has 2 heads. The character recognition accuracy of the VCRec (Ours), is significantly better than that of the other two-stage methods, which proves that the spatial structure of the character is very important. Our preliminary exploration of modeling the spatial structure of characters has achieved good results. Compared with the temporal model, VCRec (Ours) has a 4.92\% performance improvement.

Moreover, we have designed three different spatio-temporal feature fusion strategies, depicted in Fig.~\ref{fig:featurefusionmode} as (B), (C), and (D). We conduct the relevant experiments for these feature fusion strategies on AWCV-100K-UCAS2024, and the results are shown in Table~\ref{table:FeatureFusionMode}. Among them, comparing A, B, C and D have better accuracy, which shows the importance of modeling character structures. Comparing B, C and D behave better, which shows that higher-level features are more effective when modeling character structures by spatial feature encoder. Compared with C and D, D achieves the SOTA, which shows that it is more effective to apply structural information directly to stroke features. 

\begin{figure}
\begin{center}
 \includegraphics[width=1\linewidth]{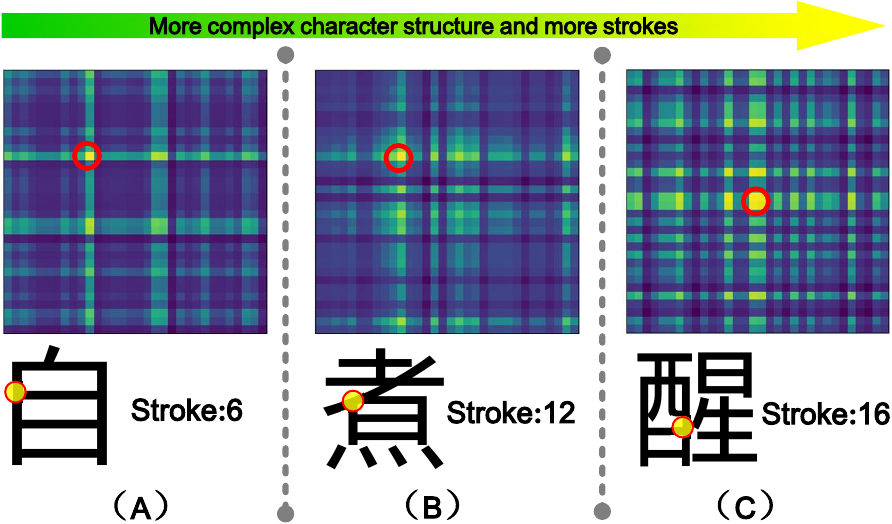}
\end{center}
   \caption{
The Visualization of StrokeGAT. (A), (B) and (C) represent three different characters. From left to right, the character structure becomes more complex and the number of strokes increases. \textbf{(TOP)} The visualization of character features through the StrokeGAT, the brighter the color, the closer the dependency between the features. \textbf{(BOTTOM)} Character and its stroke count, where the circle part corresponds to the highlighted part in the visualization.
   }
\label{fig:characterstructure}
\end{figure}

\noindent\textbf{Visual Analysis of VCRec.}
The analysis of Fig.~\ref{fig:characterstructure} unveils a compelling relationship between the intricacy of characters and the corresponding response within the StrokeGAT feature map. As depicted in the lower section of the figure, characters exhibiting higher complexity appear to induce a more pronounced and intensified response in the StrokeGAT feature map located in the upper part.

This correlation signifies StrokeGAT's remarkable ability to capture and represent the intricate web of stroke connections inherent in various characters. The heightened highlighting observed in the feature map suggests that StrokeGAT excels in discerning and emphasizing these intricate stroke patterns, providing a deeper insight into the structural composition of characters.

\subsection{Ablation Study}

\noindent\textbf{Comparison of Different Temporal Feature Encoder.} We conducted ablation studies on the Temporal Feature Encoder, as detailed in Table~\ref{table:TemporalFeatureEncoder}. In the 4th row, we observed a significant decline in performance when the Temporal Feature Encoder was not utilized, resulting in a 12.76\% lower performance compared to our method. Comparing the experimental results in rows 1-3 and our method, it is evident that 1DCNN excels in modeling temporal sequences.

\begin{table}
\centering
\caption{Results of Different Temporal Feature Encoder in VCRec on AWCV-100K-UCAS2024.}
\label{table:TemporalFeatureEncoder}
\begin{small}
\resizebox{\linewidth}{!}
{
\begin{tabular}{llll}
\toprule
\multicolumn{1}{l}{\textbf{Temporal Feature Encoder}} &\textbf{Spatial Feature Encoder} & \textbf{AR(\%) $\uparrow$} & \textbf{Params(M) $\downarrow$} \\ \midrule
\textbf{1D-TCRN~\cite{gan2020air}}        & StrokeGAT    & 49.31  &8.6 \\
\textbf{LSTM~\cite{LSTM}}    & StrokeGAT   & 47.22 & 9.5 \\
\textbf{Transformer~\cite{gan2019new}}    & StrokeGAT   & 44.11  &39.2 \\
\textbf{\XSolidBrush}  & StrokeGAT  & 39.67 & 2.9 \\ \midrule
\textbf{1DCNN(Ours)}   &  StrokeGAT & \textbf{52.43} & 3.9 \\ \bottomrule
\end{tabular}
}

\end{small}
\end{table}

\noindent\textbf{Character Structure Model.} 
In our quest to address sparse visual features, our focus delved into character spatial structures, leading to the proposition of the Spatial Feature Encoder strokeGAT. Given the significant strides of graph convolutional networks (GCN) in handling diverse unstructured data, our study ventured into character spatial exploration using GCN, SparseGAT~\cite{velickovic2017graph}, and StrokeGAT. As elucidated in Table~\ref{table:CharacterStructureModel}, StrokeGAT emerged as the optimal performer, highlighting the Graph Attention Network's (GAT) adeptness in capturing extensive spatial dependencies crucial for character analysis. This reinforces GAT's superiority in discerning and modeling intricate spatial structures pivotal for character feature understanding.

\noindent\textbf{Comparison of Different Decoders.}
In our analysis of AWCV-100K-UCAS2024, we scrutinized various decoder models. The Connectionist Temporal Classification (CTC) technique, pioneered by Graves et al.~\cite{graves2006connectionist}, empowers models to learn direct mappings from input to output sequences sans explicit alignment requirements. Our experimentation, outlined in Table~\ref{table:decoder}, encompassed exploring diverse configurations of fully connected decoders. Notably, the 2-layer fully connected (2-layer FC) decoder emerged as the best performer. 

In contrast to CTC, the 2-layer FC decoder exhibited superior performance, particularly excelling in single-character recognition tasks. This disparity highlights the distinct advantage of the FC decoder architecture over CTC, especially in accurately identifying individual characters within sequences.

\subsection{Performance on other Dataset.} 
\begin{table}
\centering
\caption{Results of Different Character Structure Model in VCRec on AWCV-100K-UCAS2024.}
\label{table:CharacterStructureModel}
\begin{small}
\begin{tabular}{lll}
\toprule
\multicolumn{1}{l}{\textbf{Character Structure Model}} & \textbf{AR(\%) $\uparrow$} & \textbf{Params(M) $\downarrow$} \\ \midrule
\textbf{GCN~\cite{kipf2016semi}}         & 51.88  &3.8 \\
\textbf{SparseGAT~\cite{velickovic2017graph}}   & 49.34 & 4.1 \\ \midrule
\textbf{StrokeGAT(Ours)}    & \textbf{52.43} & 3.9 \\ \bottomrule
\end{tabular}
\end{small}
\end{table}

\begin{table}
\centering
\caption{Results of Different Decoders in VCRec on AWCV-100K-UCAS2024.}
\label{table:decoder}
\begin{small}
\begin{tabular}{lll}
\toprule
\multicolumn{1}{l}{\textbf{Decoder}} & \textbf{AR(\%) $\uparrow$} & \textbf{Params(M) $\downarrow$} \\ \midrule
\textbf{CTC~\cite{graves2006connectionist}}   & 49.66 & 3.9  \\ 
\textbf{1-layer FC}    & 50.43 &  3.8  \\ 
\textbf{3-layer FC}    & 51.43 &  3.9  \\ \midrule
\textbf{2-layer FC(Ours)}    & \textbf{52.43} & 3.9  \\ \bottomrule
\end{tabular}
\end{small}
\end{table}

We construct different experiments on different languages ($e.g.$, English and Chinese) and forms ($e.g.$, trajectory-based and video-based) of datasets.

Table~\ref{table:TrajChinaDataset} displays the performance of VCRec on trajectory-based datasets, which have exhibited even better performance on the IAHCC-UCAS2016 (a trajectory-based Chinese character dataset). 

\begin{table}
\caption{VCRec performs on the trajectory-based Chinese character dataset IAHCC-UCAS2016~\cite{xu2015recognition}.}
\label{table:TrajChinaDataset}
\begin{small}
\begin{tabular}{lll}
\toprule
\textbf{Trajectory-based Dataset}                    & \textbf{Method}                       & \textbf{AR(\%) $\uparrow$} \\ \midrule
                                                     & \textbf{LSTM}~\cite{7903730}          & 93.18   \\
                                                     & \textbf{ViT-B}~\cite{Vit}             & 93.85   \\
                                                     & \textbf{ViT-L}~\cite{Vit}             & 93.91   \\
                                                     & \textbf{1DCNN}~\cite{gan2019new}      & 96.78   \\\cmidrule{2-3}
 \multirow{-5}{*}{\textbf{IAHCC-UCAS2016}}           & \textbf{VCRec (Ours)}        & \textbf{96.85}   \\ \bottomrule
\end{tabular}
\end{small}

\end{table}

\begin{table}
\centering
\caption{VCRec performs on video-based English datasets. WiTA~\cite{kim2022writing} is a video-based English dataset. CER is character error rate.}
\label{table:VideoEnglishDataset}
\begin{small}
\begin{tabular}{lll}
\toprule
\textbf{Video-based Dataset}                                          & \textbf{Method}           & \textbf{CER(\%) $\downarrow$}                 \\ \midrule
                                                                              & \textbf{ST-rMC}~\cite{kim2022writing}          & 92.94\\
                                                                             & \textbf{ST-R(2+1)D}~\cite{kim2022writing}      & 87.51\\
                                                                              & \textbf{ST-R3D}~\cite{kim2022writing}          & 29.24\\
                                                                             & \textbf{TR-AWR}~\cite{tan2023end}              & 29.86 \\ \cmidrule{2-3}
      \multirow{-5}{*}{\textbf{WiTA}}        & \textbf{VCRec+CTC}                          & 30.12 \\ \bottomrule
\end{tabular}
\end{small}
\end{table}

\begin{table}
\caption{VCRec performs on trajectory-based English datasets. IAHEW-UCAS2016~\cite{2019In} is a trajectory-based English dataset. CAR is character accuracy rate.}
\label{table:TrajEnglishDataset}
\resizebox{\linewidth}{!}{
\begin{tabular}{lll}
\toprule
\textbf{Tarjectory-based Dataset}                                           & \textbf{Method}           & \textbf{CAR(\%) $\uparrow$}                 \\ \midrule
                                                                             & \textbf{LSTM+CTC}~\cite{7903730}          &97.13   \\
                                                                           & \textbf{LSTM+Decoder}~\cite{bahdanau2016neural}          &96.86   \\
                                                                             & \textbf{1DCNN+Decoder}~\cite{gan2019new}        &97.45 \\\cmidrule{2-3}
 \multirow{-5}{*}{\textbf{IAHEW-UCAS2016}}        & \textbf{VCRec (Ours)}          &\textbf{96.51}   \\ \midrule
\end{tabular}}
\end{table}

Since the WiTA consists of English words containing multiple English characters, we have adopted the concept of CTC~\cite{graves2006connectionist} in those experiments instead of the FC decoder mentioned in our method. Table~\ref{table:VideoEnglishDataset} displays the performance of VCRec on a video-based English dataset, which has demonstrated comparable performance on the WiTA (English). Table~\ref{table:TrajEnglishDataset} displays the performance of VCRec on trajectory-based datasets, which have exhibited comparable performance on the IAHEW-UCAS2016 (a trajectory-based English word dataset). 

The results of the experiments on these English datasets have demonstrated comparable performance in other languages, indicating that our method exhibits strong generalization across different languages. In addition, we have proceeded to analyze that our approach showcases considerable effectiveness in modeling spatial structures within the context of the Chinese language, leveraging its inherent structural traits. Nevertheless, its performance exhibits a relatively diminished impact in English contexts, primarily due to the heightened emphasis on temporal information within the language.

\section{Conclusion}

In this work, we have presented the AWCV-100K-UCAS2024 dataset, a video-based air-writing dataset designed for real-world scenarios. To address the challenges posed by AWCV-100K-UCAS2024, we propose the VCRec method, a two-stage architecture. This method initially compresses sparse visual features into fingertip features and then models fingertip feature sequences using a spatial-temporal sequence module. This module captures temporal information from fingertip movements and represents the spatial structure of Chinese characters. VCRec achieves an accuracy of 52.43\% on the AWCV-100K-UCAS2024 dataset. 

We anticipate that our dataset and the baseline model will stimulate research in real-world applications. For example, applying air-writing in healthcare for hands-free control in sterile environments, utilizing it in AR and VR for immersive experiences, integrating it into education and training scenarios, enhancing accessibility for individuals with disabilities, and implementing gesture-based interfaces in industrial settings for improved safety and efficiency in manufacturing processes.

The dataset, toolkit, and experimental results will be released to further advance air-writing research.


\begin{IEEEbiography}[{\includegraphics[width=1in,height=1.25in,clip,keepaspectratio]{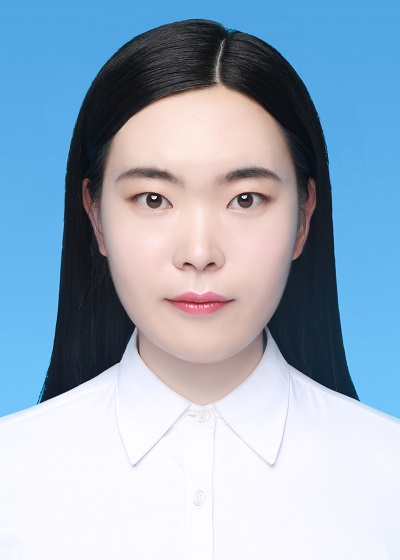}}]{Meiqi Wu}
received the MSc Degree from the University of Science and Technology of China In 2021 and continued to study for her doctorate at the School of Computer Science and Technology, University of Chinese Academy of Sciences, Beijing, China, in the same year. Her current research interests include computer vision, human-computer interaction, and machine learning.
\end{IEEEbiography}

\begin{IEEEbiography}[{\includegraphics[width=1in,height=1.25in,clip,keepaspectratio]{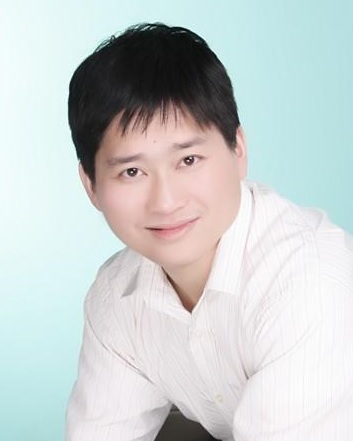}}]{Kaiqi Huang} received the BSc and MSc degrees from the Nanjing University of Science Technology, China, and the PhD degree from Southeast University. He is currently a full professor with the Center for Research on Intelligent System and Engineering, Institute of Automation, Chinese Academy of Sciences. He is also with the University of Chinese Academy of Sciences (UCAS), and the CAS Center for Excellence in Brain Science and Intelligence Technology. He has authored or co-authored more than 210 papers in important international journals and conferences, such as the IEEE TPAMI, IJCV, T-IP, T-SMCB, TCSVT, Pattern Recognition, CVIU, ICCV, ECCV, CVPR, ICIP, and ICPR. His current research interests include computer vision, pattern recognition, and game theory, including object recognition, video analysis, and visual surveillance. He is the co-chair and program committee member of more than 40 international conferences, such as ICCV, CVPR, ECCV, and the IEEE workshops on visual surveillance. He is an associate editor for IEEE Transactions on Systems, Man, and Cybernetics: Systems and Pattern Recognition.
\end{IEEEbiography}

\begin{IEEEbiography}[{\includegraphics[width=1in,height=1.25in,clip,keepaspectratio]{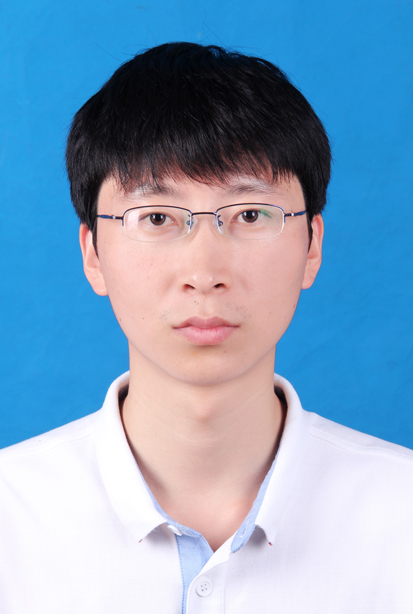}}]{Yuanqiang Cai} 
received the Ph.D. degree from the University of Chinese Academy of Sciences, Beijing, China, in 2021. He is currently a lecturer at the Beijing University of Posts and Telecommunications. His research interests include object detection, multimedia content analysis, and text localization and recognition in images and videos. He has published more than 10 papers in referred conferences and journals including NeurIPS, AAAI, ACM MM, TCSVT, and PR.
\end{IEEEbiography}

\begin{IEEEbiography}[{\includegraphics[width=1in,height=1.25in,clip,keepaspectratio]{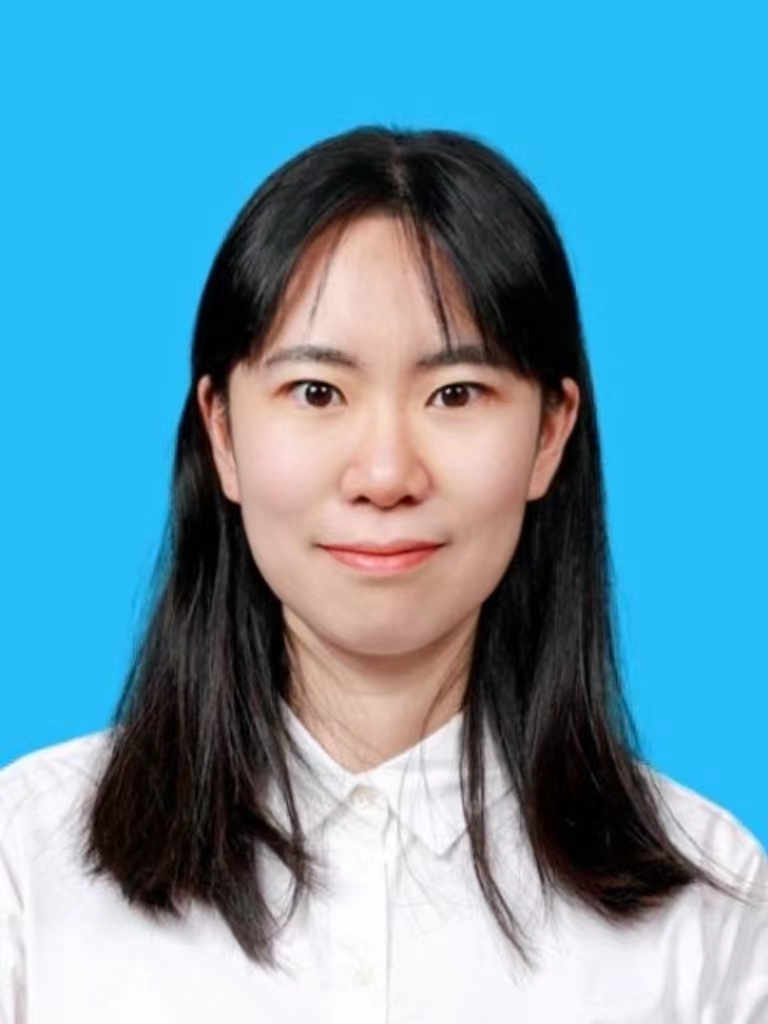}}]{Shiyu Hu}, Ph.D., Institute of Automation, Chinese Academy of Sciences. Shiyu Hu received her PhD degree from the University of Chinese Academy of Sciences in Jan. 2024. She has authored or coauthored more than 10 research papers in the areas of computer vision and pattern recognition at international journals and conferences, including TPAMI, IJCV, NeurIPS, etc. Her research interests include computer vision, visual object tracking, and visual intelligence evaluation.
\end{IEEEbiography}

\begin{IEEEbiography}[{\includegraphics[width=1in,height=1.25in,clip,keepaspectratio]{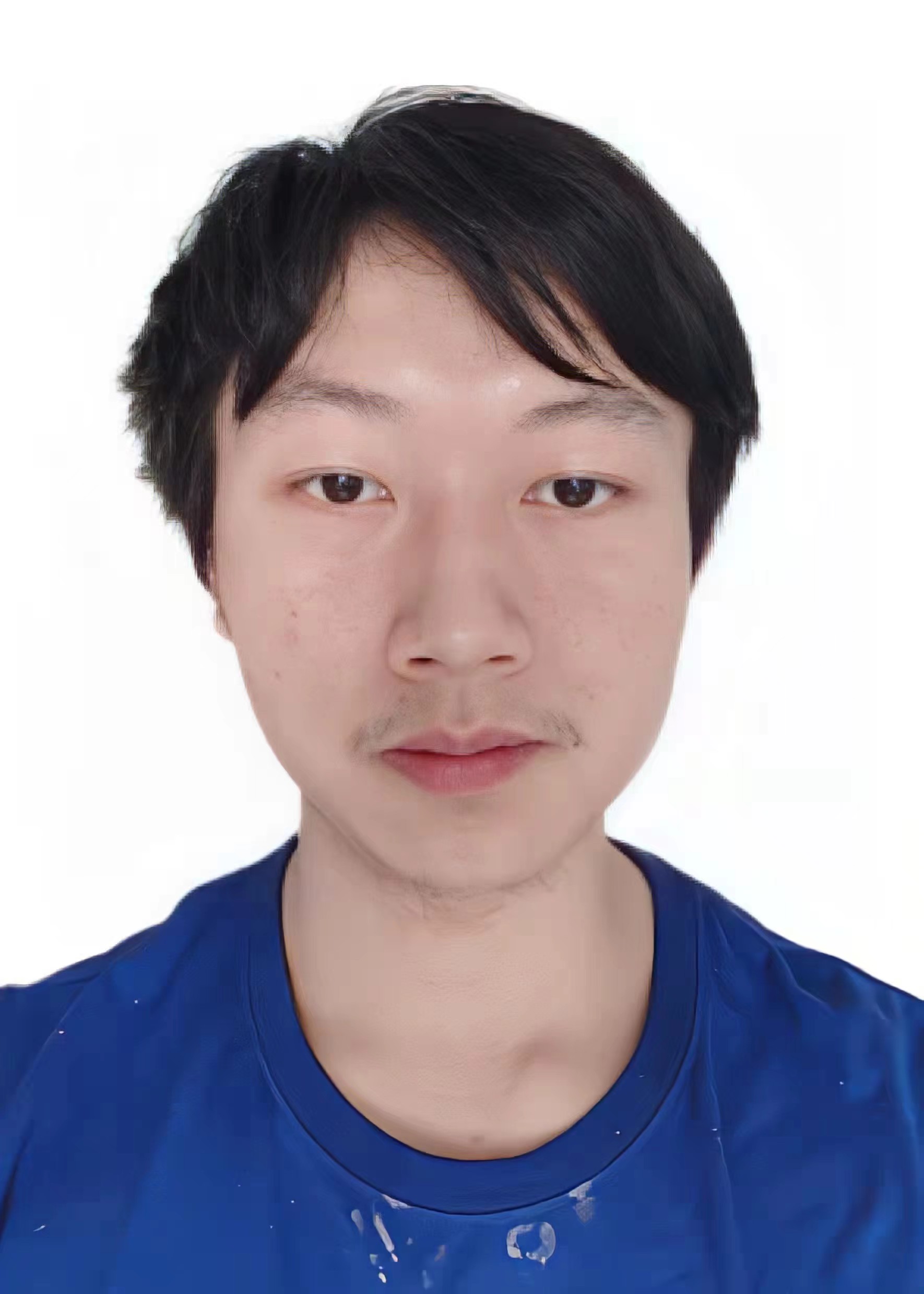}}]{Yuzhong Zhao}
received a B.S. degree from Peking University, Beijing, China in 2020, and he is currently pursuing an M.E. degree at the University of Chinese Academy of Science, Beijing. His research interests include computer vision, scene text detection, and recognition.
\end{IEEEbiography}

\begin{IEEEbiography}[{\includegraphics[width=1in,height=1.25in,clip,keepaspectratio]{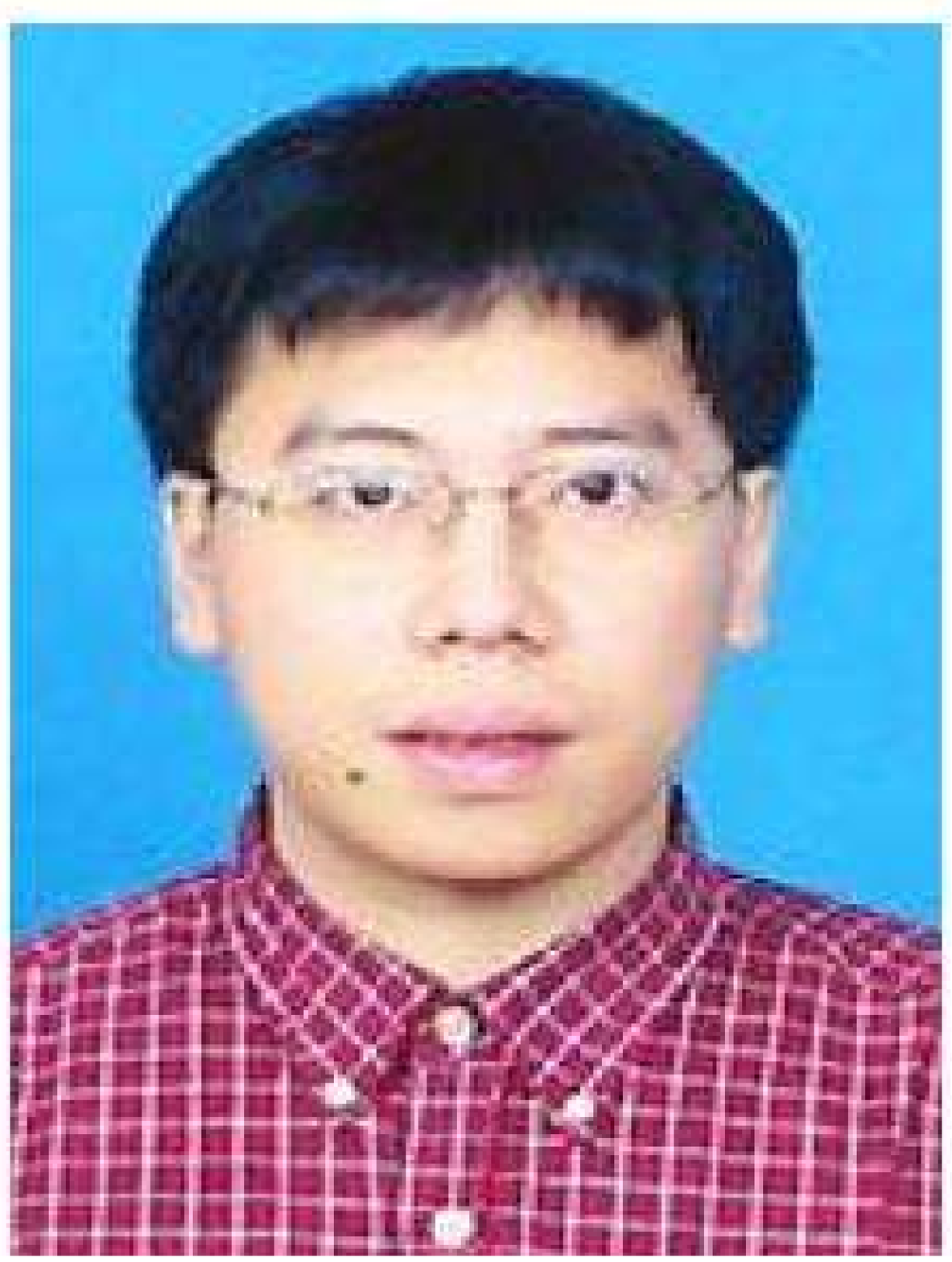}}]{Weiqiang Wang} 
received the B.E. and M.S. degrees in
computer science from Harbin Engineering University, in 1995 and 1998, respectively, and the Ph.D. degree in computer science from the Institute of Computing Technology (ICT), Chinese Academy of Sciences (CAS), China, in 2001.
He is currently a Professor at the School of Computer Science and Technology, University of Sciences. His research interests include multimedia content analysis, computer vision, pattern recognition, and human-computer interaction.
\end{IEEEbiography}

\end{document}